\RequirePackage{fix-cm}
\documentclass[final, 3p,twocolumn]{elsarticle}

\usepackage{siunitx}
\DeclareSIUnit\px{px} % Defines \px to display as "px"

\usepackage{amsmath}
\usepackage{caption}
\usepackage{subcaption}
\usepackage{tikz}
\usepackage{xurl} % Allow long documentation URLs to wrap within columns.
\usepackage{adjustbox}
\usepackage{booktabs}
\usepackage{graphicx} % for \resizebox
\usepackage{adjustbox} % <-- robust alternative to \resizebox

\usepackage{rotating}
\newcommand{\tableimage}[1]{%
	\vspace{2mm}\raisebox{-0.5\height}{\includegraphics[width=\linewidth,height=0.16\textheight,keepaspectratio]{#1}}\vspace{2mm}%
}

\usepackage{tikz}
\usetikzlibrary{arrows.meta,positioning,fit,calc,shapes.geometric, shadows, backgrounds, shapes.misc, decorations.pathmorphing}

\tikzset{
  >=Latex,
  textnode/.style={align=left, font=\small},
  stepbox/.style={draw, rounded corners, very thick, fill=white},
  arrow/.style={-Latex, very thick},
  icon/.style={line width=0.7pt, draw=black},
  thinicon/.style={line width=0.5pt, draw=black},
  every node/.style={font=\footnotesize},
}

\definecolor{std}{HTML}{2F5597}   % standard layers (blue)
\definecolor{mod}{HTML}{D35400}   % modified layers (orange)
\definecolor{graytext}{HTML}{555555}

\tikzset{
  >={Stealth[length=3mm]},
  block/.style={draw, rounded corners=2pt, very thick, align=center,
                minimum width=3.2cm, minimum height=1.15cm, fill=white},
  stdblock/.style={block, draw=std, text=std},
  modblock/.style={block, draw=mod, text=mod},
  smallnote/.style={font=\footnotesize\itshape, text=graytext},
  arrow/.style={-Stealth, very thick},
  legendbox/.style={draw, rounded corners=2pt, inner sep=4pt}
}

\everymath=\expandafter{\the\everymath\displaystyle}

\definecolor{SERgreen}{HTML}{1ff01f}
\definecolor{SERgray}{HTML}{909090}
\definecolor{SERwhite}{HTML}{ffffff}
\definecolor{SERblue}{HTML}{80b3ff}

\DeclareSIUnit{\angstrom}{\text{\AA}}

\usepackage{graphicx}% Include figure files
\usepackage{dcolumn}% Align table columns on decimal point
\usepackage[utf8]{inputenc}
\usepackage[T1]{fontenc}
\usepackage{newtxtext}
\usepackage{newtxmath}
\usepackage{bm}% Load after the math fonts for correct bold symbols.
\usepackage{microtype}
\usepackage{ragged2e}
\usepackage{etoolbox}
\usepackage[hidelinks]{hyperref}
\begin{document}
\raggedbottom % Avoid stretching sparse, figure-heavy columns vertically.

\begin{frontmatter}

\title{Deep Learning Quantification of Bainite Fractions from Optical Micrographs of Bearing Steel}

\author{Sebasti\'an~Echeverri Restrepo\corref{corresponding}}
\ead{sebastian.echeverri.restrepo@skf.com}
\author{Stephan~Bagger{\"o}hr}
\author{Hanzheng~Huang}
\author{Diana~Airas~Prol}
\author{Sandra~Bonjer}
\author{Stefan~Geerts}
\author{Boris~Minov}
\author{Cees~Taal}
\author{Pei~Yan}

\address{SKF Research \& Technology Development (RTD), SKF B.V., Meidoornkade 14, 3992 AE Houten, The Netherlands}

\cortext[corresponding]{Corresponding author.}

\begin{abstract}
Quantifying the bainite fraction in mixed bainite--martensite microstructures in bearing {steels is} important for process and quality control {in bearing manufacture}. In such steels, bainite and martensite exhibit strongly overlapping features in light-optical micrographs, often precluding reliable manual assessment. We present a deep learning regression method that predicts a dilatometry-derived reference bainite fraction directly from unsegmented optical images. Reference bainite fractions spanning the transformation range were generated by interrupting the bainitic transformation at points identified from the dilatometer length-change response. A deep neural network was trained using the   captured optical micrographs and corresponding dilatometry-derived bainite fractions. The validation of the trained network covered  held-out specimens within the same microscope domain, repeated metallographic preparation, and images from two additional microscopes. The lowest prediction errors were obtained within the original microscope domain while maintaining strong agreement after repeated metallographic preparation. Larger errors were observed for images acquired with the additional microscopes. {The iterated mean filter} reduced prediction errors for the two additional imaging workflows. These results demonstrate the feasibility of predicting reference bainite fractions directly from unsegmented optical images of high-carbon bearing steel. The approach requires neither image segmentation nor manual phase annotation.

\end{abstract}


\end{frontmatter}

%%%%%%%%%%%%%%%%%%%%%%%%%%%%%%%%%%%%%%%%%%%%%%%%
%%%%%%%%%%%%%%%%%%%%%%%%%%%%%%%%%%%%%%%%%%%%%%%%
%%%%%%%%%%%%%%%%%%%%%%%%%%%%%%%%%%%%%%%%%%%%%%%%

\section{Introduction}

Bainitically hardened bearing steels are commonly used for large-section bearings because they combine strength, toughness, rolling-contact fatigue resistance, and comparatively low heat-treatment distortion~\cite{Bhadeshia2012,Lai2016a,Amey2012}. A mixed bainite--martensite microstructure offers a route to shortening the isothermal hold while targeting a similar balance of properties. This microstructure can be produced by interrupting the bainitic transformation before completion and then quenching to transform part of the remaining austenite into martensite. Quantifying the resulting constituent fractions is therefore important for heat-treatment development and quality control. {In industrial heat treatment, differences in component geometry, furnace loading, and local heat transfer can produce variations in thermal history and final microstructure, even within a controlled process. Quantification from metallographic images therefore provides a complementary means of verifying the resulting bainite fraction for quality control.}

Distinguishing bainite from martensite in these mixed microstructures becomes increasingly difficult as their relative fractions become comparable. In high-carbon steel, {fine microstructural features can be difficult to distinguish by light microscopy~\cite{Holm2021}}. {Lath bainite and martensite can exhibit similar optical morphologies despite differences in their lattice defect structures and composition~\cite{Zisman2015}. Conventional nital and picral etching can therefore leave constituent identification ambiguous~\cite{Hasbi2021}. Although colour etching can improve differentiation, the observed colours and the reproducibility of phase quantification depend on the reagent and etching conditions~\cite{Hasbi2021,Xiao2025}. Specimen preparation and imaging conditions introduce additional variability~\cite{Bachmann2026}. Consequently, manual quantification of the bainite fraction from optical images can be subjective and difficult to reproduce~\cite{Bachmann2023}, motivating methods that estimate the overall fraction without requiring unambiguous local identification of each constituent.}

Previous machine-learning studies have focused mainly on identifying and separating individual phases in steels with lower carbon contents than the grade investigated here. Most formulate the problem as classification or segmentation and rely on detailed phase labels, often obtained using EBSD or other complementary characterisation methods. In~\cite{Muller2021}, previously segmented carbon-rich regions in scanning-electron micrographs were classified using textural and morphological features. In~\cite{Bachmann2023}, patch classification was combined with sliding-window segmentation of optical and scanning-electron micrographs supported by correlative EBSD. The approaches in~\cite{Breumier2022} and~\cite{Bachmann2024} used EBSD data either as the model input or to generate masks for training optical-image segmentation.

More recent studies have followed a similar direction. In~\cite{Hallo2026}, expert-defined regions, unsupervised segmentation, and machine-learning classification were combined, while~\cite{Bachmann2026} examined transfer between microscopes using annotated optical-image patches. Complementing these image-based studies,~\cite{Xiao2025} compared X-ray diffraction, metallography, dilatometry, and EBSD for bainite--martensite quantification in a medium-carbon steel.

These studies demonstrate substantial progress, but the reviewed machine-learning tasks are formulated as classification or segmentation and depend on local phase assignments, selected regions, constituent boundaries, or pixel-wise masks. This leaves direct regression of a continuous bainite fraction from unsegmented light-optical images as an open task. {For bearing heat-treatment development and quality control, the overall bainite fraction is often the quantity of interest, rather than its detailed spatial distribution. Direct regression avoids the need to delineate individual constituents or construct pixel-wise training masks.}

Evidence for cross-microscope performance is also limited for high-carbon steels, in which optical differentiation of bainite and martensite is particularly difficult. Differences in illumination, optics, acquisition software, export settings, and specimen preparation can introduce domain shift, making transferability across imaging workflows an important evaluation requirement.

In this work, we develop a modified ResNet-18~\cite{He2016} model for regression of bainite fraction from optical micrographs of one high-carbon bearing-steel grade using specimen-level labels. The training and inference procedures require neither prior segmentation, pixel- or region-level manual annotation, nor EBSD maps. Reference bainite fractions are derived from dilatometric transformation progress, thereby avoiding inter-observer variability in expert phase annotation. The validation of the model covers interpolation within one microscope domain, repeated metallographic preparation, and acquisition with two additional microscopes. {An iterated mean filter with smoothing parameter $N$} is also evaluated as a preprocessing strategy for these imaging workflows. This design provides a direct optical regression workflow and an assessment of sensitivity to imaging conditions and software settings.

%%%%%%%%%%%%%%%%%%%%%%%%%%%%%%%%%%%%%%%%%%%%%%%%
%%%%%%%%%%%%%%%%%%%%%%%%%%%%%%%%%%%%%%%%%%%%%%%%
%%%%%%%%%%%%%%%%%%%%%%%%%%%%%%%%%%%%%%%%%%%%%%%%

\section{Materials and experimental methods}
\label{sec:production}
\subsection{Material}

A high-carbon bearing steel, 100CrMnMoSi8-4-6, was investigated in this work. Its initial spheroidised condition consists primarily of ferrite and cementite. Table~\ref{table-a} presents its nominal chemical composition.

\begin{table}[ht]
    \centering
    \resizebox{\columnwidth}{!}{%
    \begin{tabular}{lccccc}
        \hline
        Steel & C & Si & Mn & Cr & Mo \\
        \hline
        100CrMnMoSi8-4-6 & 1.0 & 0.5 & 0.9 & 2.0 & 0.5 \\
        \hline
    \end{tabular}
    }
    \caption{Nominal chemical composition of the investigated steel (wt\%).}
    \label{table-a}
\end{table}

\subsection{Heat treatment and reference bainite fractions}
\label{subsec:heat-treatment}

\begin{figure}[!t]
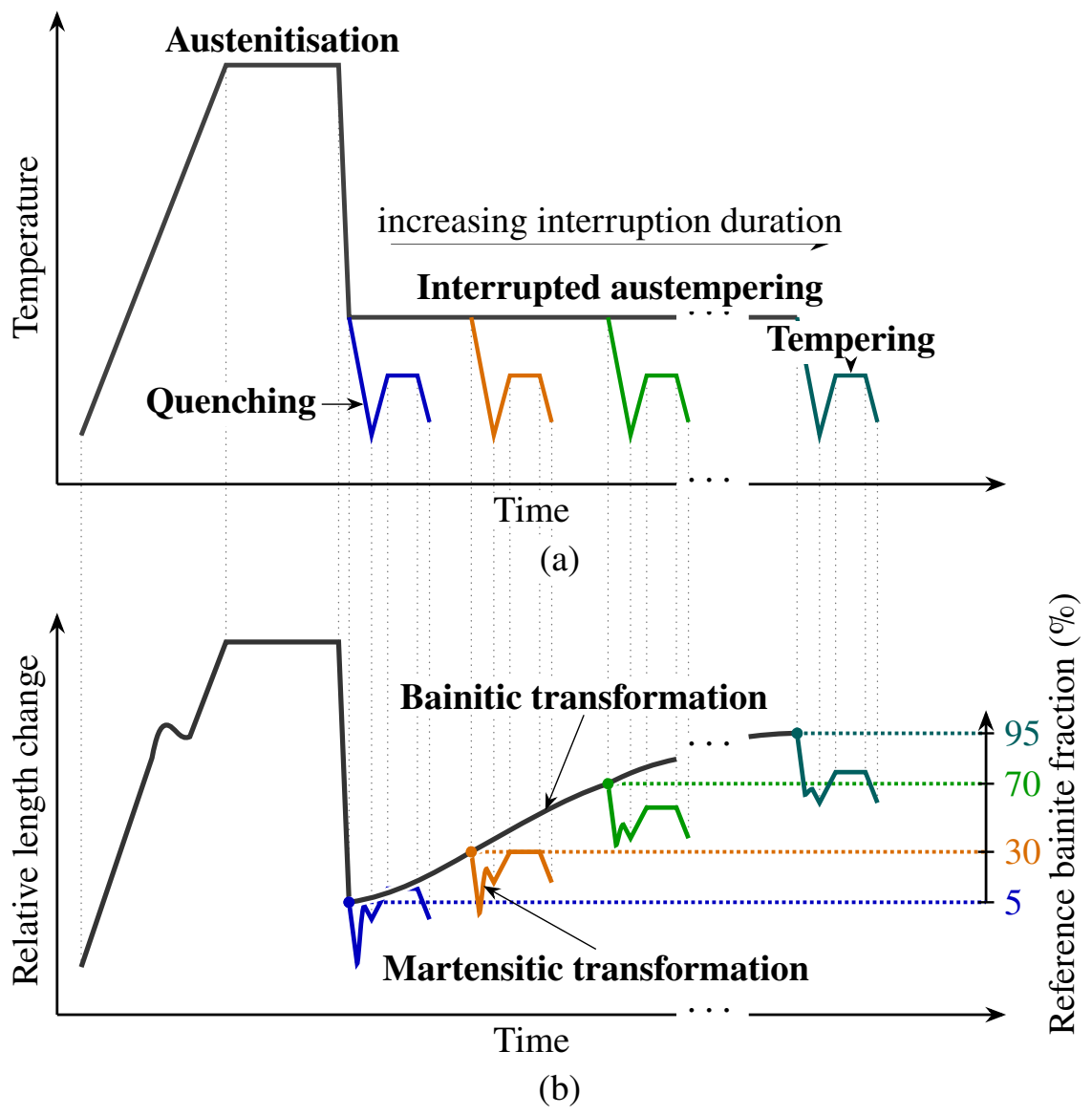

  \centering
  \begin{subfigure}[t]{\columnwidth}
    \centering
    \resizebox{\columnwidth}{!}{% A shared canvas lets stage guides join the two panels at identical times.
% Both source panels have x=0.85 cm and the bounding box (0,-0.35)-(13.6,6.75).
% Their vertical units are 0.82 cm in (a) and 0.72 cm in (b).
\begingroup
\setbox0=\hbox{\input{Figures/Figure1_martensitic_expansion/heat_treatment_overview}}
\setbox2=\hbox{\input{Figures/Figure1_martensitic_expansion/label_generation_overview}}
\begin{tikzpicture}[
  x=0.85cm, y=1cm,
  stageguide/.style={draw=black!55, line width=0.40pt,
    dash pattern=on 0pt off 1.8pt, line cap=round}
]
  % Draw behind the panels and their labels. Entries are time / temperature
  % panel height / length-response panel height at each thermal-stage boundary.
  \foreach \stagetime/\temperatureheight/\lengthheight in {
    0.85/0.78/1.05,
    2.65/5.55/5.82,
    4.05/5.55/5.82,
    4.18/2.30/2.00,
    4.46/0.78/1.75,
    4.66/1.55/2.20,
    5.03/1.55/2.20,
    5.18/0.95/1.75,
    5.70/2.30/2.74,
    5.98/0.78/2.29,
    6.18/1.55/2.74,
    6.55/1.55/2.74,
    6.70/0.95/2.29,
    7.40/2.30/3.74,
    7.68/0.78/2.94,
    7.88/1.55/3.39,
    8.25/1.55/3.39,
    8.40/0.95/2.94,
    9.75/2.30/4.48,
    10.03/0.78/3.46,
    10.23/1.55/3.91,
    10.60/1.55/3.91,
    10.75/0.95/3.46
  } {
    \draw[stageguide]
      (\stagetime,{5.70+(\temperatureheight+0.35)*0.82})
      -- (\stagetime,{(\lengthheight+0.35)*0.72});
  }
  \node[anchor=south west, inner sep=0pt, outer sep=0pt]
    at (0,5.70) {\box0};
  \node[anchor=south west, inner sep=0pt, outer sep=0pt]
    at (0,0) {\box2};
  \node[font=\fontsize{11}{13}\selectfont, fill=white, inner sep=2pt]
    at (6.80,5.30) {(a)};
  \node[font=\fontsize{11}{13}\selectfont, fill=white, inner sep=2pt]
    at (6.80,-0.30) {(b)};
\end{tikzpicture}
\endgroup}
    % Panel labels are drawn on the shared canvas; retain subfigure references.
    \phantomcaption
    \label{fig:heattreatment}
    \phantomcaption
    \label{fig:label_generation}
  \end{subfigure}

  \caption{Schematic of the heat-treatment route and reference-bainite-fraction selection. (a) Temperature--time paths used to generate mixed bainite--martensite specimens. (b) Relative-length response with representative interruption levels between \SI{5}{\percent} and \SI{95}{\percent}; the normalisation endpoints are \SI{0}{\percent} and \SI{100}{\percent}. The corresponding coloured branches show quenching, tempering, and the final quench.}
  \label{fig:thermal_and_dilatometry}
\end{figure}

Cylindrical dilatometry specimens with a diameter of \SI{4}{\milli\metre} and a length of \SI{10}{\milli\metre} were prepared. The specimens were heat treated in a TA Instruments DIL 805 quenching dilatometer. Each cycle comprised austenitisation, controlled cooling, an interrupted isothermal transformation, quenching, and tempering, as summarised schematically in Figure~\ref{fig:heattreatment}.

Individual dilatometry specimens were heat treated at each of nineteen conditions spanning reference bainite fractions from \SI{5}{\percent} to \SI{95}{\percent} in increments of \SI{5}{\percent}, with the isothermal holding time varied to obtain the target fractions. The relative-length response $d(t)$ was linearly normalised between the start and endpoint of a completed reference bainitic transformation to define the reference bainite fraction,

\begin{equation}
  f_{\mathrm{B}}^{\mathrm{ref}}(t)
  = 100\,\frac{d(t)-d_{\mathrm{s}}}{d_{\mathrm{e}}-d_{\mathrm{s}}},
  \label{eq:dilatometry_fraction}
\end{equation}

where $d_{\mathrm{s}}$ and $d_{\mathrm{e}}$ are the relative-length values at the start and end of the reference bainitic transformation, respectively, and $f_{\mathrm{B}}^{\mathrm{ref}}$ is expressed in percent. These endpoints were assigned reference fractions of \SI{0}{\percent} and \SI{100}{\percent}. This relation assumes that the relative-length change during the isothermal transformation is proportional to bainitic transformation progress. The linear normalisation applies to the relative-length response; it does not assume a constant transformation rate. Interruption times for the imaged specimens were selected where the reference curve reached the target fractions, as illustrated in Figure~\ref{fig:label_generation}.

\subsection{Metallographic preparation}

% Before submission, confirm ASTM E3 compliance and edition and verify the final abrasive size against the preparation record.
Metallographic preparation followed ASTM E3, with final polishing using a \SI{1}{\micro\metre} abrasive. {After heat treatment, the cylindrical specimens were ground longitudinally to produce a flat surface on one side. Each specimen was then embedded in resin with this flat surface exposed. The exposed surface was polished to remove a layer approximately \SI{500}{\micro\metre} thick. Finally, the surface was etched with \SI{1.5}{\percent} nital (nitric acid in ethanol) to reveal the microstructure for light-optical microscopy.}

\section{Imaging and datasets}
\label{sec:datasets}

Five image datasets were acquired from the specimens described in Section~\ref{subsec:heat-treatment} using three microscopes, $\text{M}_1$, $\text{M}_2$ and $\text{M}_3$ (Table~\ref{tab:datasets}). Datasets from $\text{M}_1$ were acquired automatically with a Zeiss Axio Observer~7 microscope using an oil-immersion objective. Images from $\text{M}_1$ were exported in ZEN core version 3.5.96.03000 using the ``Apply display settings and channel color'' option, incorporating the configured display settings into the exported pixel values. Dataset $\text{M}_2$ was acquired using a second Zeiss Axio Observer~7 microscope with a non-oil-immersion objective. Dataset $\text{M}_3$ was acquired manually using an Olympus K-201 microscope, also without oil immersion.

All images were acquired at a nominal magnification of $1000\times$. The $\text{M}_1$ superscripts identify the training, validation, and {repeated metallographic preparation (``Reprep'')} datasets; $\text{M}_2$ and $\text{M}_3$ denote the collections from the other microscopes. Specimens are the independent material units, while images and sub-images are repeated observations within specimens. Table~\ref{tab:bainite_images2} presents representative exported micrographs from the three microscopes at selected reference bainite fractions before preprocessing.

\begin{table*}[t]
    \centering
    \caption{Overview of the datasets used in this work.}
    \label{tab:datasets}
    \begin{tabular}{lccccc}
        \toprule
        Dataset & Microscope & Acquisition mode & Bainite fraction (\%) & Images per fraction & Image size\\
        \midrule

        $\text{M}_1^\text{Train}$ & Zeiss~1 & Automatic &
        \begin{tabular}{@{}c@{}}
            \numlist{5;10;15;25;30}\\
            \numlist{35;45;50;55;65}\\
            \numlist{70;75;85;90;95}
        \end{tabular}
        & 144 & $2752\times2208$ \\[4pt]

        $\text{M}_1^\text{Validation}$ & Zeiss~1 & Automatic &
        \numlist{20;40;60;80}
        & 144 & $2752\times2208$ \\[4pt]

        $\text{M}_1^\text{Reprep}$ & Zeiss~1 & Automatic &
        \begin{tabular}{@{}c@{}}
            \numlist{10;20;30;40;50}\\
            \numlist{60;70;80;90}
        \end{tabular}
        & 144 & $2752\times2208$ \\[4pt]

        $\text{M}_2$ & Zeiss~2 & Automatic &
        \begin{tabular}{@{}c@{}}
            \numlist{5;15;25;35;45}\\
            \numlist{55;65;75;85;95}
        \end{tabular}
        & 144 & $2752\times2208$ \\[4pt]

        $\text{M}_3$ & Olympus & Manual &
        \begin{tabular}{@{}c@{}} 
            \numlist{10;20;30;40;50}\\
            \numlist{60;70;80;90} 
        \end{tabular}
        & 20 & $1920\times1080$\\

        \bottomrule
    \end{tabular}
\end{table*}

\newsavebox{\tablemicrographbox}
\begin{table*}[p]
    \centering
    \caption{Sample exported micrographs before preprocessing at different dilatometry-derived reference bainite fractions for microscopes $\text{M}_1$, $\text{M}_2$, and $\text{M}_3$. The $\text{M}_1$ examples are from the training and same-microscope validation datasets. Dashes indicate fractions not represented in the corresponding dataset.}
    \label{tab:bainite_images2}
    \small
    \setlength{\tabcolsep}{4pt}
    % The reference 10-micrometre bar spans approximately 220 original pixels.
    % Argument 2 is the original image width before resizing to 1000 pixels.
    \renewcommand{\tableimage}[2]{%
        \sbox{\tablemicrographbox}{\includegraphics[width=\linewidth,height=0.13\textheight,keepaspectratio]{#1}}%
        \pgfmathsetlengthmacro{\tablebarlength}{(220/#2)*\wd\tablemicrographbox}%
        \begin{tikzpicture}[baseline=(micrograph.center)]
            \node[inner sep=0pt,outer sep=0pt] (micrograph) {\usebox{\tablemicrographbox}};
            \node[anchor=south east,inner sep=1.3pt,outer sep=0pt,fill=white]
                at ([xshift=-2pt,yshift=2pt]micrograph.south east) {%
                \begin{tikzpicture}
                    \draw[black,line width=0.6pt] (0pt,0pt) -- (\tablebarlength,0pt);
                    \draw[black,line width=0.6pt] (0pt,-1.1pt) -- (0pt,1.1pt);
                    \draw[black,line width=0.6pt] (\tablebarlength,-1.1pt) -- (\tablebarlength,1.1pt);
                    \node[anchor=south,inner sep=0pt,outer sep=0pt,font=\normalfont\scriptsize]
                        at (0.5*\tablebarlength,2pt) {\SI{10}{\micro\metre}};
                \end{tikzpicture}%
            };
        \end{tikzpicture}%
    }
    \begin{tabular}{@{}>{\centering\arraybackslash}m{0.16\textwidth}*{3}{>{\centering\arraybackslash}m{0.26\textwidth}}@{}}
        \toprule
        \shortstack{Reference bainite\\fraction (\%)} & $\text{M}_1$ & $\text{M}_2$ & $\text{M}_3$ \\
        \midrule
        10 & \tableimage{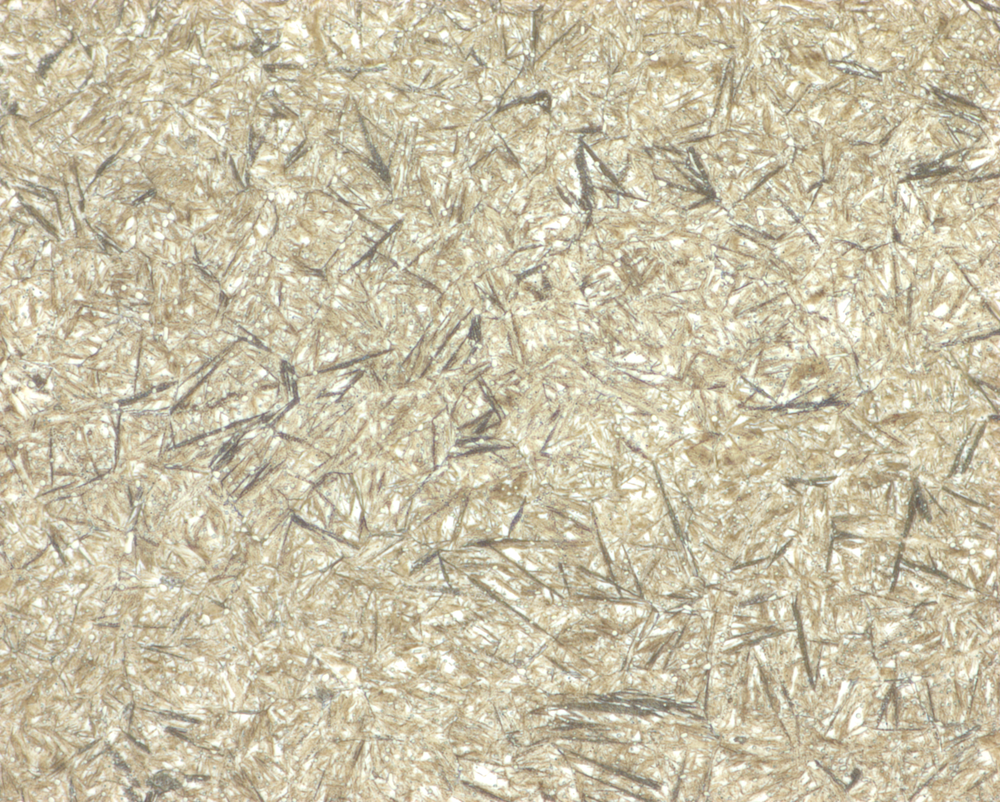}{2752} & --- & \tableimage{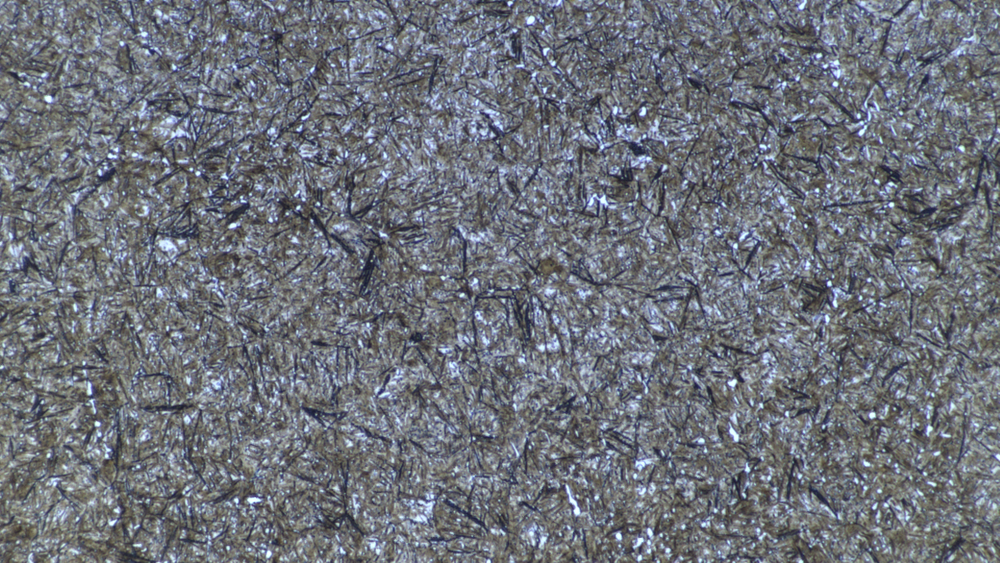}{1920} \\
        \addlinespace[4pt]
        25 & \tableimage{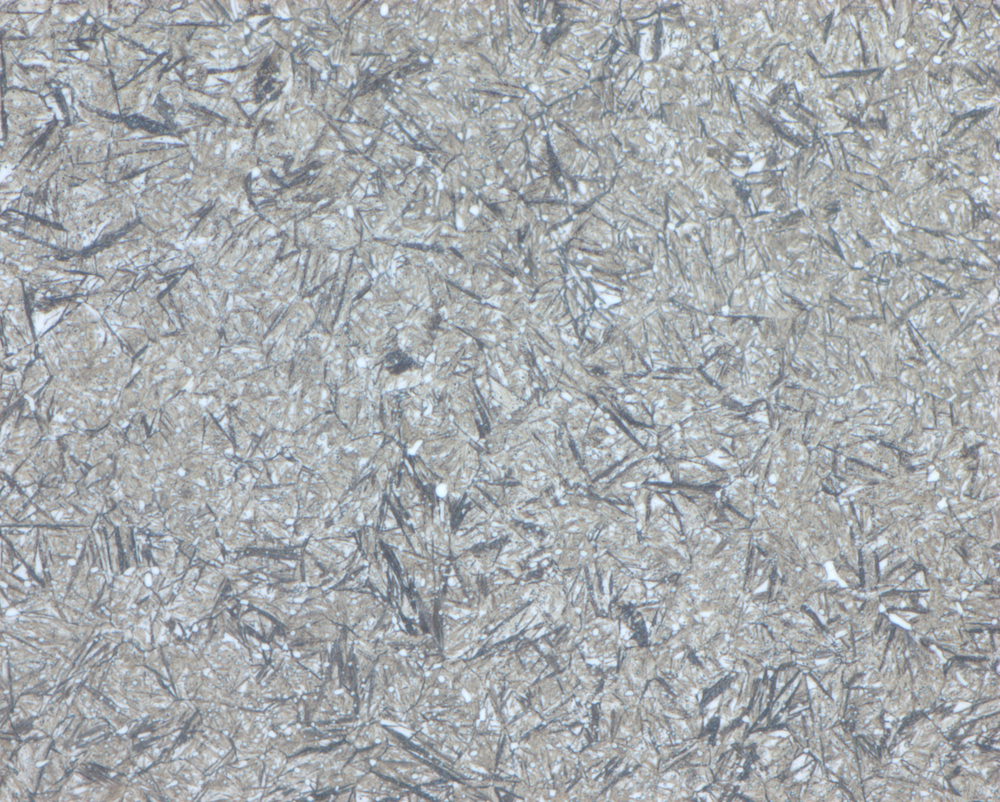}{2752} & \tableimage{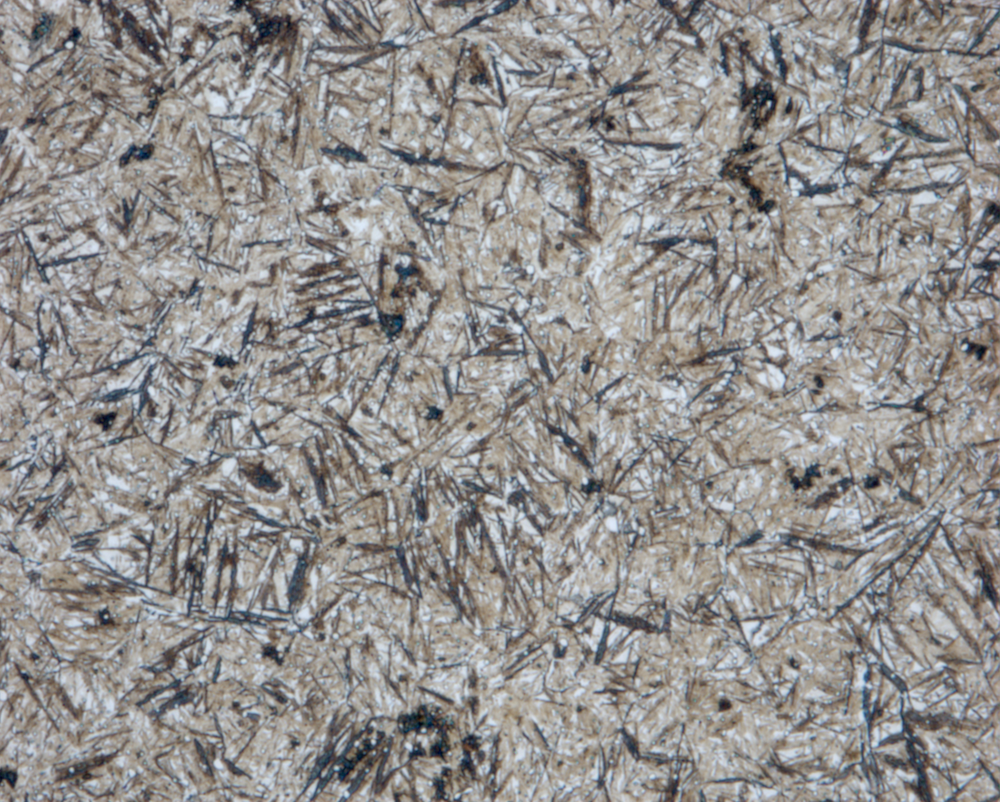}{2752} & --- \\
        \addlinespace[4pt]
        40 & \tableimage{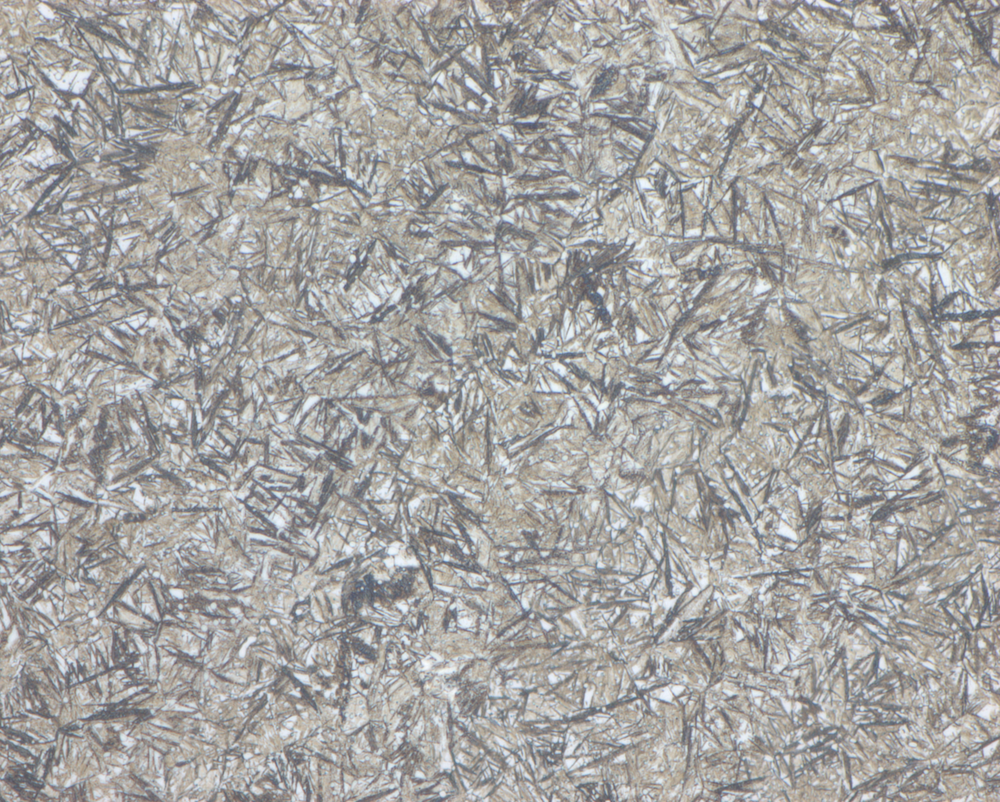}{2752} & --- & \tableimage{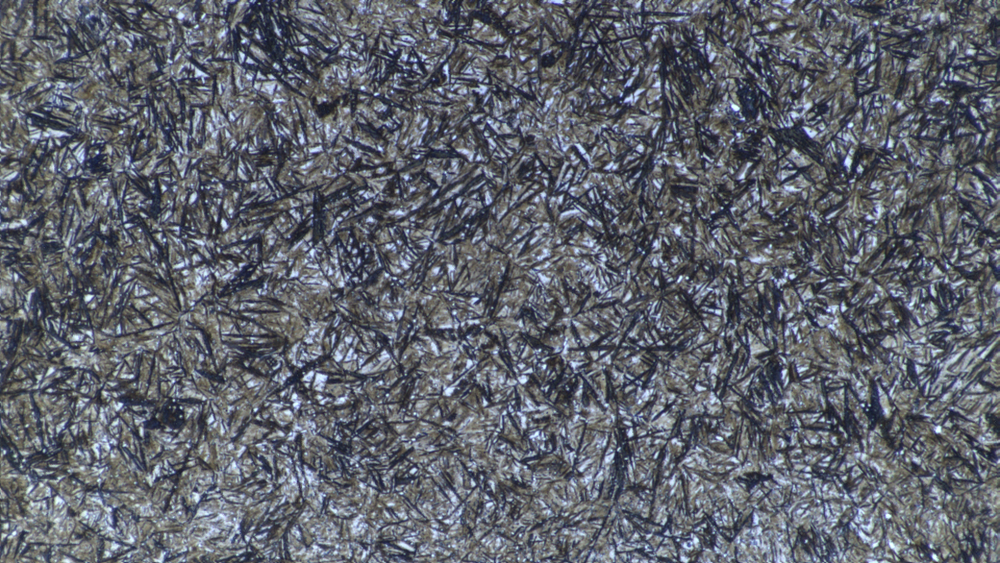}{1920} \\
        \addlinespace[4pt]
        55 & \tableimage{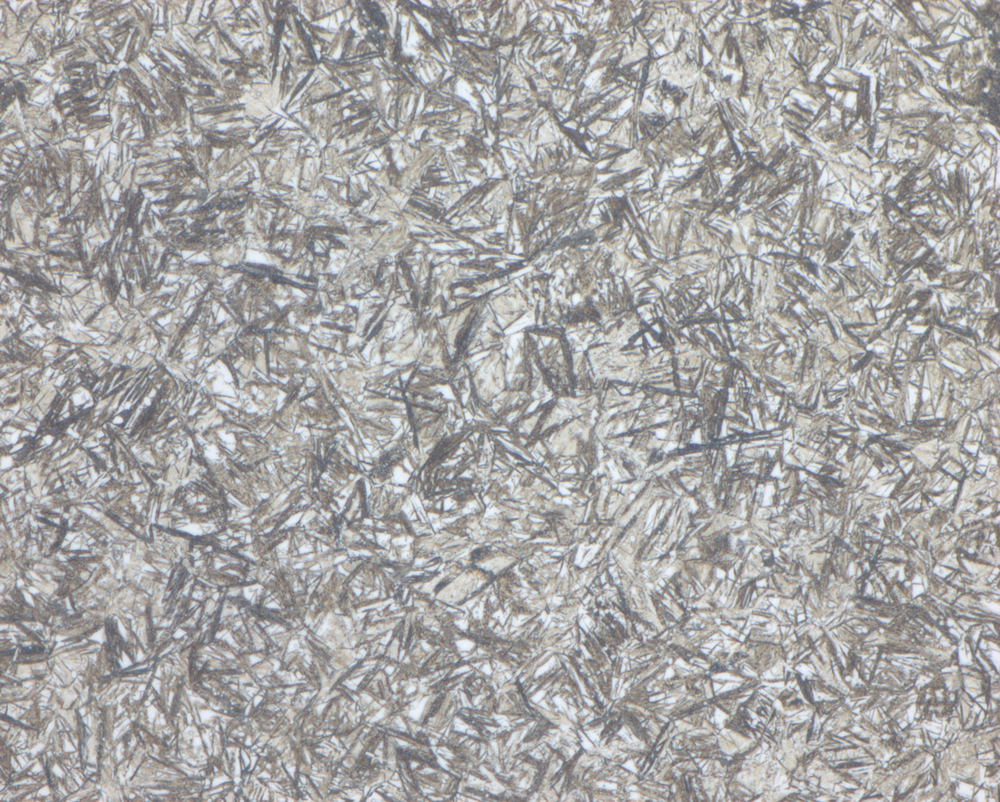}{2752} & \tableimage{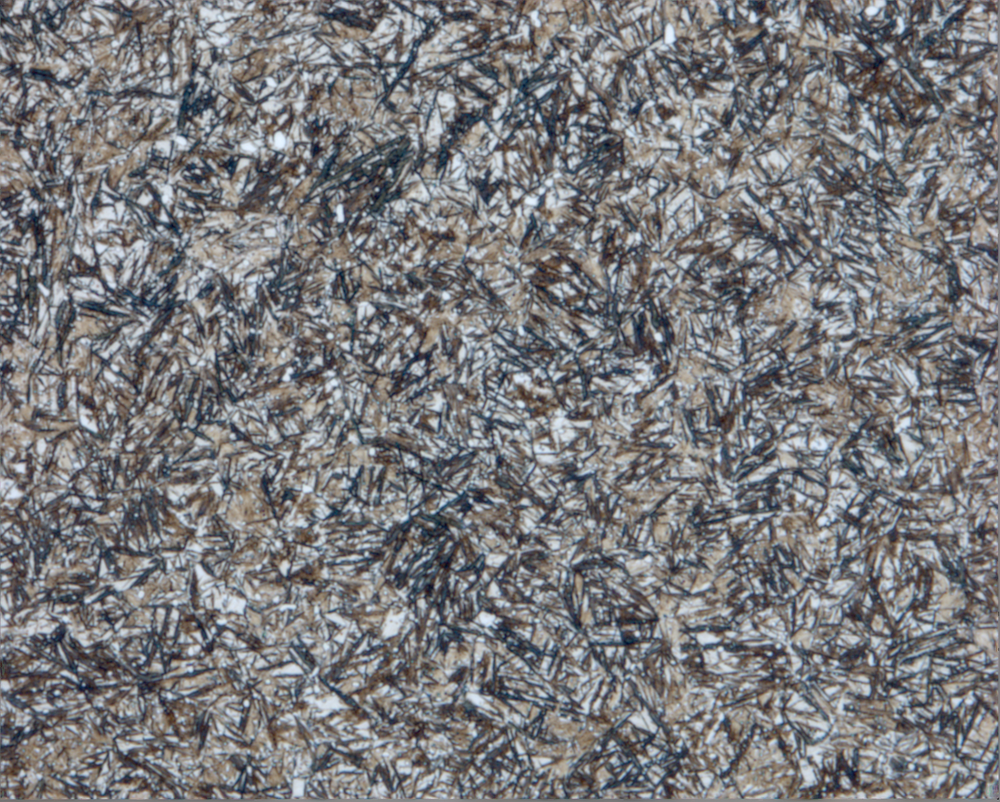}{2752} & --- \\
        \addlinespace[4pt]
        70 & \tableimage{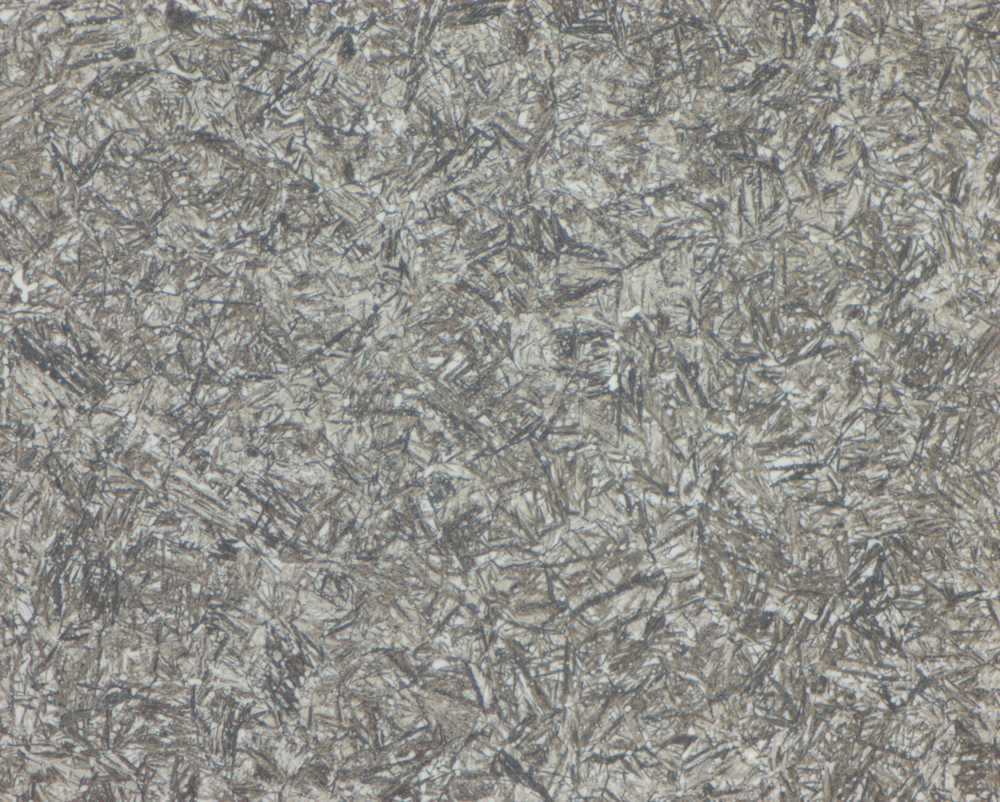}{2752} & --- & \tableimage{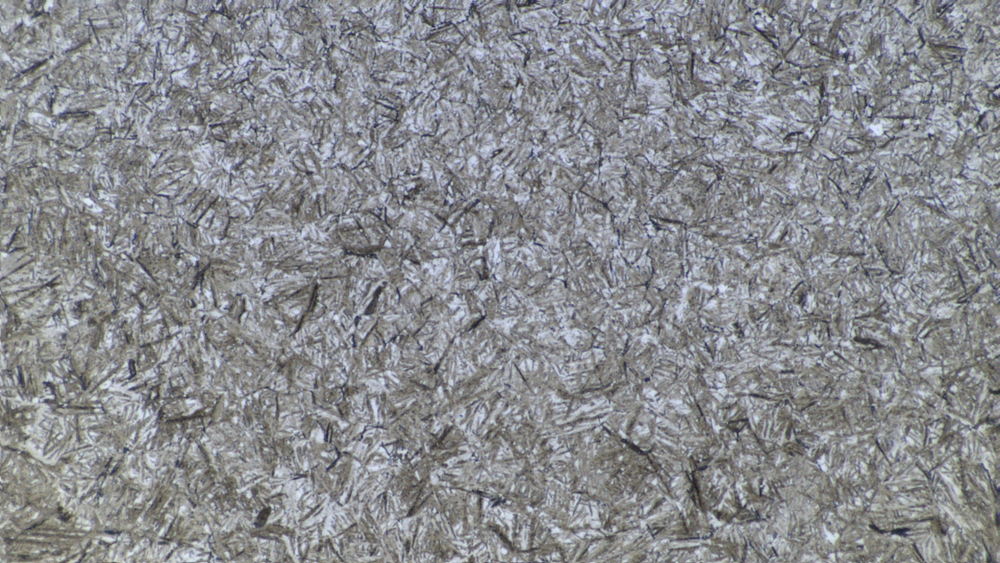}{1920} \\
        \addlinespace[4pt]
        85 & \tableimage{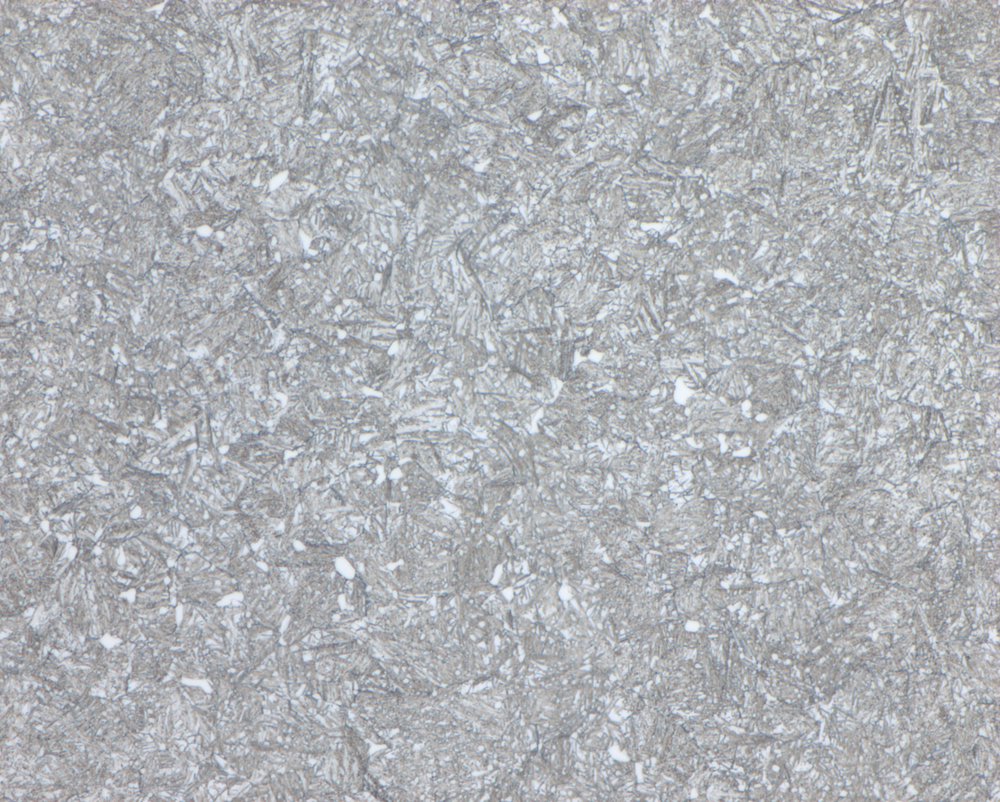}{2752} & \tableimage{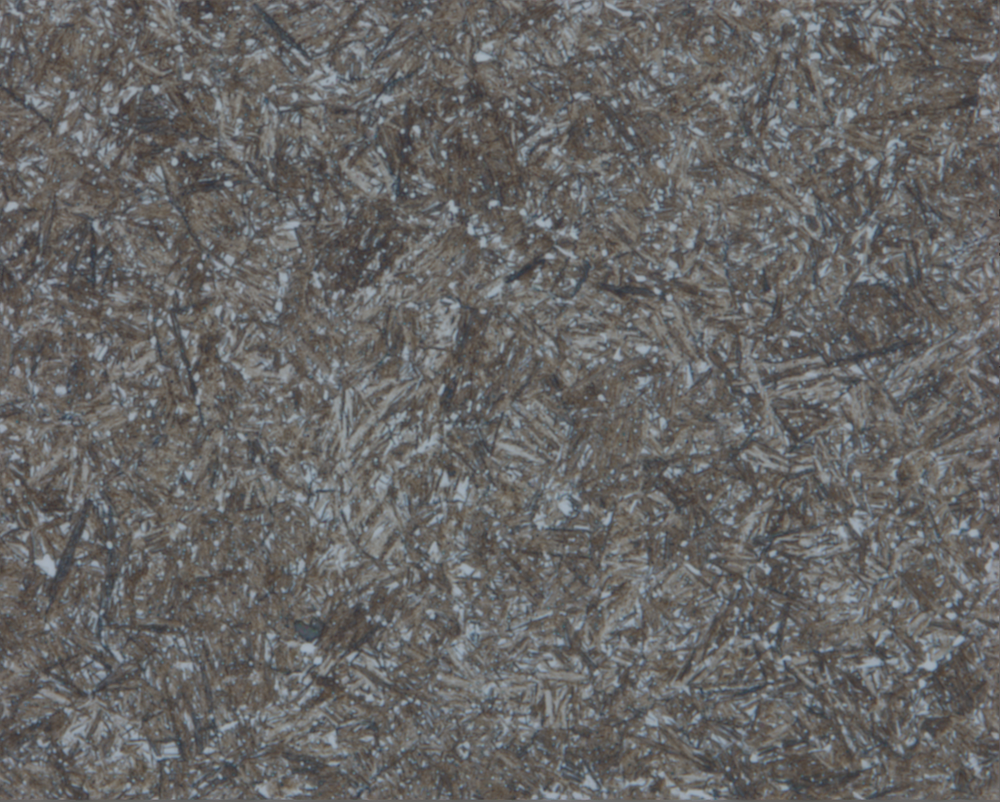}{2752} & --- \\
        \bottomrule
    \end{tabular}
\end{table*}

Fifteen specimens supplied the training data. The four specimens at reference bainite fractions of \SIlist{20;40;60;80}{\percent} supplied the validation data, and all images from these specimens were excluded from training. This split evaluates interpolation to intermediate fractions on held-out specimens within the same microscope domain.

For $\text{M}_1^\text{Reprep}$, previously imaged specimens were ground, polished, and etched again, and different surface locations were sampled using the same microscope. This collection evaluates the effect of repeated preparation and spatial sampling.

To evaluate transfer between microscopes, additional images of the same specimen series were acquired with $\text{M}_2$ at reference bainite fractions from \SI{5}{\percent} to \SI{95}{\percent} and with $\text{M}_3$ from \SI{10}{\percent} to \SI{90}{\percent}, in steps of \SI{10}{\percent} for both datasets. {Metallographic preparation was repeated before acquiring the $\text{M}_2$ and $\text{M}_3$ images.}

\section{Regression workflow}

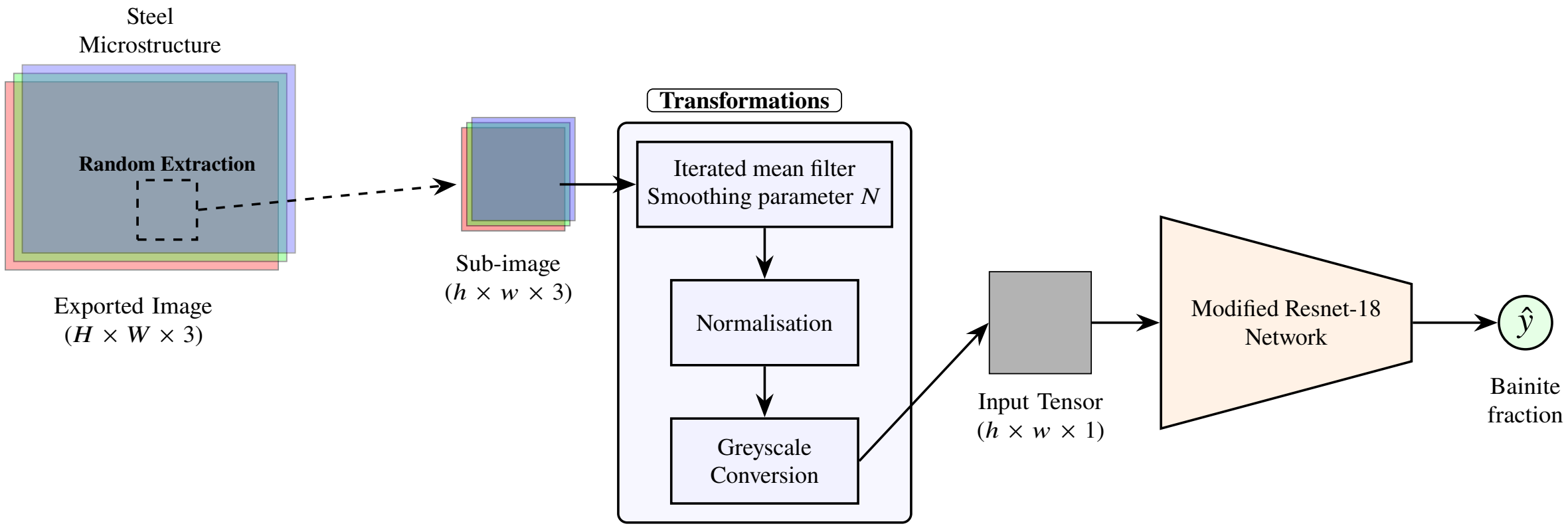
\begin{figure*}[t]
    \begin{adjustbox}{max width=\textwidth}%
    \begin{tikzpicture}[
    node distance=1.2cm and 1.0cm,
    font=\sffamily\small,
    % Styles
    box/.style={draw, rectangle, rounded corners=2pt, fill=white, minimum width=2cm, minimum height=1cm, align=center, drop shadow={opacity=0.1}},
    proc/.style={draw, rectangle, fill=blue!5, minimum width=2.2cm, minimum height=1cm, align=center, thick},
    nn/.style={draw, trapezium, trapezium angle=75, fill=orange!10, minimum width=2.5cm, minimum height=1.5cm, align=center, thick, shape border rotate=270},
    arrow/.style={-{Stealth[scale=1.2]}, thick},
    label_text/.style={text width=2.5cm, align=center, font=\footnotesize}
]

    % 1. Exported Image Representation  (NOW A RECTANGLE)
    \node (orig_img) [draw=none, fill=none, minimum width=3.2cm, minimum height=2.2cm] at (0,0) {};
    \node [align=center] at ($(orig_img.north)+(0.2,0.71cm)$) {Steel \\ Microstructure};
    \node [below=0.1cm of orig_img, label_text] {Exported Image \\ ($H \times W \times 3$)};

    % Layered channels visualization (RGB only)
    \draw[fill=red!70,   opacity=0.45] ($(orig_img.north west)+(0.10,0.10)$) rectangle ($(orig_img.south east)+(0.10,0.10)$);
    \draw[fill=green!70, opacity=0.40] ($(orig_img.north west)+(0.20,0.20)$) rectangle ($(orig_img.south east)+(0.20,0.20)$);
    \draw[fill=blue!70,  opacity=0.35] ($(orig_img.north west)+(0.30,0.30)$) rectangle ($(orig_img.south east)+(0.30,0.30)$);

% 2. Random sub-image extraction
    \node (crop_rect) [draw, dashed, black, thick, minimum width=0.7cm, minimum height=0.7cm] at ($(orig_img.center) + (0.4,-0.3)$) {};
\node [black, scale=0.7, above=0.01cm of crop_rect, font=\bfseries] {Random Extraction};

% 3. Sub-image
    \node (sub_img) [draw=none , fill=none, minimum width=1.2cm, minimum height=1.2cm, right=2.2cm of orig_img] {};
\node [below=0.1cm of sub_img, label_text] {Sub-image \\ ($h \times w \times 3$)};
    \draw [arrow, dashed, black] (crop_rect.east) -- (sub_img.west);

% RGB channel overlays for the sub-image (RGB only)
    \draw[fill=red!75,   opacity=0.50] ($(sub_img.north west)+(0.06,0.06)$) rectangle ($(sub_img.south east)+(0.06,0.06)$);
    \draw[fill=green!75, opacity=0.45] ($(sub_img.north west)+(0.12,0.12)$) rectangle ($(sub_img.south east)+(0.12,0.12)$);
    \draw[fill=blue!75,  opacity=0.40] ($(sub_img.north west)+(0.18,0.18)$) rectangle ($(sub_img.south east)+(0.18,0.18)$);

    % 4. Transformations — vertical flow
    \node (avg)  [proc, right=0.9cm of sub_img] {Iterated mean filter \\ Smoothing parameter $N$};
    \node (norm) [proc, below=6mm of avg] {Normalisation};
    \node (gray) [proc, below=6mm of norm] {Greyscale \\ Conversion};

    % Draw a global block BEHIND the three transforms
    \begin{pgfonlayer}{background}
        \node (tf_block) [
            draw, rounded corners=3pt, thick, inner sep=6pt,
            fill=blue!3,
            fit = (avg) (norm) (gray)
        ] {};
        % Block title as a small tab at the top-left of the block
        \node[font=\bfseries\footnotesize, fill=white, draw, rounded corners=2pt, inner xsep=4pt, inner ysep=1pt]
            at ($(tf_block.north west)+(1.5cm, 0.25cm)$) {Transformations};
    \end{pgfonlayer}

    % Connecting inside the block
    \draw [arrow] (sub_img.east) -- (avg.west);
    \draw [arrow] (avg) -- (norm);
    \draw [arrow] (norm) -- (gray);

    % 5. Neural Network Input Tensor
    \node (gs_img) [draw, fill=gray!60, minimum width=1.2cm, minimum height=1.2cm, right=0.9cm of tf_block] {};
    \node [below=0.1cm of gs_img, label_text] {Input Tensor \\ ($h \times w \times 1$)};
    \draw [arrow] (gray.east) -- (gs_img.west);

    % 6. Regression Model
    \node (network) [nn, right=0.8cm of gs_img] {Modified Resnet-18 \\ Network};
    \draw [arrow] (gs_img) -- (network);

    % 7. Final Output (Scalar)
    \node (output) [circle, draw, fill=green!10, thick, right=1.0cm of network, inner sep=2pt] {\large $\hat{y}$};
    \node [below=0.2cm of output, label_text] {Bainite \\ fraction};
    \draw [arrow] (network) -- (output);

\end{tikzpicture}
    \end{adjustbox}
    \caption{End-to-end pipeline from an optical micrograph to a predicted bainite fraction.}
    \label{fig:Training_pipeline}
\end{figure*}

The regression workflow comprises sub-image sampling, preprocessing, network training, and inference ({Figure~\ref{fig:Training_pipeline}}).

\subsection{Sub-image sampling}

For each training batch, 30 source micrographs were selected randomly with overlap from $\text{M}_1^\text{Train}$, and 15 square sub-images, each measuring $256\times256$ pixels, were sampled at random positions within each selected micrograph. With 15 reference bainite fractions in the training dataset, a batch contains two source micrographs per fraction on average. This two-stage sampling produces 450 sub-images per batch, with the number limited by the available memory. The sub-images retained their acquired orientation under the working assumption of in-plane microstructural isotropy.

\subsection{Image preprocessing}
\label{sec:preprocessing}

Each extracted sub-image was processed by {an iterated mean filter}, channel-wise standardisation, and greyscale conversion.

\begingroup

First, each colour channel $I_c$, $c \in \{R,G,B\}$, is smoothed using an iterated mean filter with smoothing parameter $N$ and normalised $3\times3$ kernel
\begin{equation}
  K = \frac{1}{8}
  \begin{bmatrix}
    1 & 1 & 1 \\
    1 & 0 & 1 \\
    1 & 1 & 1
  \end{bmatrix}.
  \label{eq:neighbour_kernel}
\end{equation}
The smoothing parameter $N$ specifies the number of iterations of this kernel, which excludes the centre pixel. The smoothed channel is
\begin{equation}
  \tilde{I}_c = K^{*N} * I_c,
  \label{eq:neighbour_average}
\end{equation}
where $*$ denotes spatial convolution and $K^{*N}$ is the $N$-fold convolution of $K$, with $K^{*0}=\delta$ denoting the identity kernel. Reflective boundary conditions are applied at the image edges for each iteration. Increasing $N$ strengthens smoothing, while $N=0$ leaves the image unchanged by this step.
\par\endgroup

Second, each channel of the smoothed sub-image is standardised individually,
\begin{equation}
  {\hat{I}_c(x,y) = \frac{\tilde{I}_c(x,y)-\mu_c}{\sigma_c}},
  \label{eq:per_image_norm}
\end{equation}
where $\mu_c$ and $\sigma_c$ are the mean and standard deviation of {the smoothed} channel $c$ computed over that sub-image alone. This per-image normalisation removes offsets in brightness and contrast. Third, the three normalised channels are converted to a single greyscale channel using the ITU--R BT.601--2 luma transform,
\begin{equation}
  \begin{aligned}
  Y(x,y) ={}& 0.2989\,\hat{I}_R(x,y)
               + 0.5870\,\hat{I}_G(x,y) \\
             &+ 0.1140\,\hat{I}_B(x,y),
  \end{aligned}
  \label{eq:greyscale_conversion}
\end{equation}
where $Y(x,y)$ is the single-channel value that forms the network input.

{The iterated mean filter was introduced to suppress the comb-like modulation observed in the intensity histograms of $\text{M}_1$ images (see Figure~\ref{fig:preprocessing_histograms}), associated with the display settings applied during export. We hypothesised that reducing this pattern would reduce the model's sensitivity to the $\text{M}_1$ export characteristics and improve generalisation to images acquired with other microscopes. The smoothing parameter $N$ was varied between model fits to evaluate this hypothesis. Channel-wise standardisation reduces differences in brightness and contrast, while greyscale conversion limits reliance on colour differences between imaging workflows.}

\subsection{Network architecture}

ResNet-18~\cite{He2016} was adapted from a classification architecture to a single-channel input and scalar regression of the normalised bainite fraction.

The network was initialised using the ImageNet-1K~\cite{Deng2009} pretrained ResNet-18 weights provided by TorchVision (\texttt{IMAGENET1K\_V1})~\cite{TorchVisionResNet18}. The initial convolution was replaced with a single-channel variant whose weights were initialised by averaging the pretrained RGB kernels over the input-channel dimension. The residual blocks and batch-normalisation layers were initialised from the pretrained model. The 1000-class fully connected classifier was replaced with a linear layer followed by a sigmoid activation, yielding a normalised scalar. All layers were trainable, and the pretrained backbone was fine-tuned jointly with the regression head.

\subsection{Training and optimisation}

{The networks were trained on a virtual machine equipped with an NVIDIA Tesla T4 GPU and 28~GB of system RAM.} Each sub-image inherited the reference bainite fraction of its source specimen, and mean squared error (MSE) between predictions and these labels was used for back-propagation.

Adam~\cite{Kingma2014} was used with an initial learning rate of \num{3e-5}, $\beta_1=0.9$, and $\beta_2=0.999$. A step scheduler reduced the learning rate by $\gamma=0.1$ every 20 epochs. {Training was terminated after 60 epochs, by which point the validation loss had largely stabilised (see Figure~\ref{fig:epoch-loss}).}

\subsection{Inference and performance evaluation}
\label{subsec:inference}

{The smoothing parameter was set to $N=15$ for comparison with $N=0$, based on the lowest logged sub-image MSE on $\text{M}_2$ alone after training. Performance on $\text{M}_3$ was then evaluated at this setting.} Predictions for {Figure~\ref{fig:bainite_predictions_panel}} and Table~\ref{tab:metrics_meanpred} were obtained using the trained model.

Each source image included in inference was sampled with 15 random sub-images. At reference fraction $k$, predictions were pooled across the sampled fields of the corresponding specimen. If ${P_k}$ denotes the number of sub-image predictions, the mean prediction is

\begin{equation}
 \bar{y}_k=\frac{1}{{P_k}}\sum_{r=1}^{{P_k}}\hat{y}_{kr}.
 \label{eq:level_mean}
\end{equation}

The evaluated predictions comprise ${P_k}=2145$ sub-images per fraction for the $\text{M}_1$ datasets, between 1635 and 2145 for $\text{M}_2$, and between 300 and 315 for $\text{M}_3$.

Performance was assessed using mean squared error (MSE), mean absolute error (MAE), and the coefficient of determination ($R^2$), calculated by comparing the mean prediction at each reference bainite fraction with its reference value. Each reference fraction received equal weight. Reference and predicted bainite fractions are expressed in \%; MAE is reported in \% and MSE in \%$^2$, while $R^2$ is dimensionless.

Performance was evaluated using one mean prediction per reference bainite fraction. The model was compared with a baseline that always predicted \SI{50}{\percent} bainite, corresponding to the mean reference fraction in the training dataset. The baseline errors depend on how far the reference fractions in each dataset lie from \SI{50}{\percent}.

\section{Results}
\subsection{Image-intensity distributions and preprocessing effects}

\begin{figure*}[!t]
  \centering
  \begingroup
  \graphicspath{{Figures/}}
  \resizebox{0.94\textwidth}{!}{\input{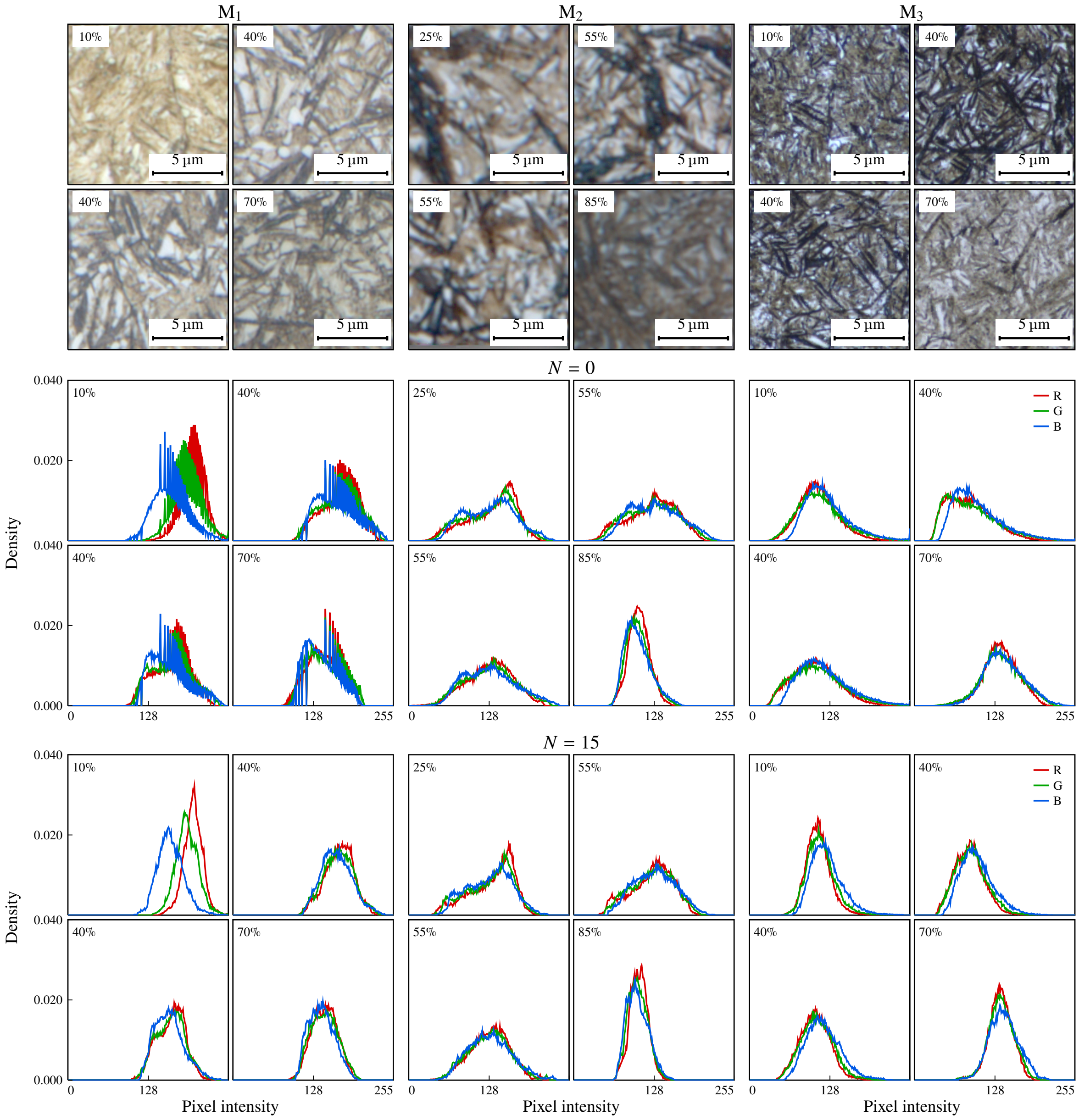}}
  \endgroup
  \caption{Illustrative effect of {the iterated mean filter} on four representative sub-images per microscope. Each pair of columns corresponds to $\text{M}_1$, $\text{M}_2$, or $\text{M}_3$. Within each pair, the $256\times256$-pixel sub-images before filtering form a $2\times2$ block at the top. The middle and bottom blocks show the corresponding RGB-channel intensity distributions {before filtering ($N=0$) and after filtering ($N=15$)}, in the same arrangement. Panel labels give the dilatometry-derived reference bainite fractions. At the patch scale shown, differences in contrast and apparent sharpness between preparation and imaging workflows illustrate the image variability that can complicate cross-microscope transferability.}
  \label{fig:preprocessing_histograms}
\end{figure*}

{Figure~\ref{fig:preprocessing_histograms}} compares four representative sub-images from each microscope and their intensity distributions before and after {application of the iterated mean filter}. The reference bainite fractions are \SIlist{10;40;70}{\percent} for $\text{M}_1$ and $\text{M}_3$, and \SIlist{25;55;85}{\percent} for $\text{M}_2$, with two different sub-images sampled at the middle reference fraction for each microscope. These examples differ in brightness, channel balance, and histogram shape. The comb-like modulation in the selected $\text{M}_1$ histograms is suppressed {by the iterated mean filter at $N=15$}, while differences in histogram location, width, and channel balance remain across the examples.

\subsection{Training and preprocessing sweep}

\begin{figure*}[htbp]
    \begin{adjustbox}{width=1.0\linewidth,center}
        \input{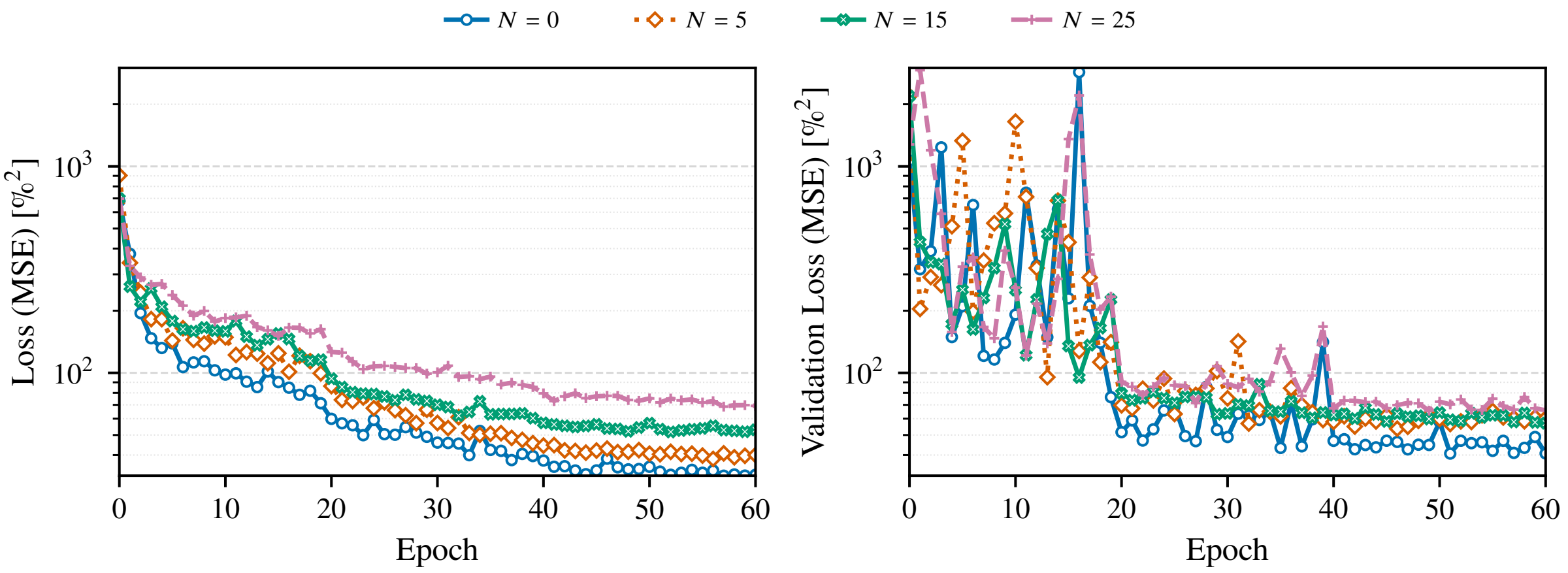}
    \end{adjustbox}
  \caption{Mean squared error versus epoch for training (left) and validation (right) datasets {using the iterated mean filter with different values of the smoothing parameter $N$}. {Only $N=0,5,15,25$ are shown; the other values are omitted for clarity.}}
  \label{fig:epoch-loss}
\end{figure*}

{Figure~\ref{fig:epoch-loss}} shows MSE over 60 epochs for the training (left) and validation (right) datasets. Within this sweep, the lowest training and validation losses occur at {$N=0$}.

\begin{figure*}[!tp]
    \centering
    \begin{adjustbox}{width=0.48\textwidth,center}
  \input{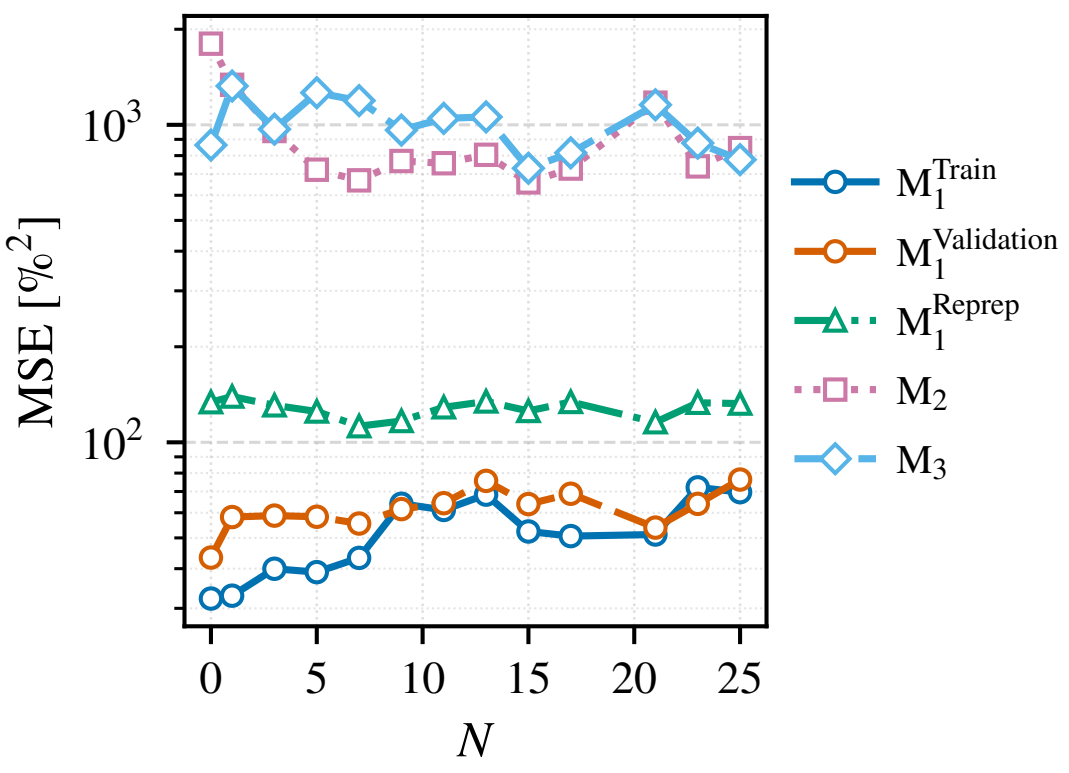}
    \end{adjustbox}
  \caption{Logged sub-image MSE after training as a function of {the smoothing parameter $N$ of the iterated mean filter} for all microscopy datasets.}
  \label{fig:errors_vs_adjustments}
\end{figure*}

{Figure~\ref{fig:errors_vs_adjustments}} shows the logged sub-image MSE  for models trained with {smoothing parameter $N$} from 0 to 25 and evaluated on each image dataset. The $\text{M}_1$ training and validation results favour {$N=0$}, while $\text{M}_1^\text{Reprep}$ changes comparatively little over the sweep. {$N=15$ was selected based on the lowest logged sub-image MSE on $\text{M}_2$. This setting also yielded the lowest MSE on $\text{M}_3$ among the evaluated values.}

\subsection{Fraction predictions across datasets}

{Figure~\ref{fig:bainite_predictions_panel}} shows the mean prediction for each reference fraction, and Table~\ref{tab:metrics_meanpred} reports the corresponding metrics. Without {the iterated mean filter} ({$N=0$}), MAE was \SI{1.52}{\percent} for $\text{M}_1^\text{Train}$ and \SI{4.13}{\percent} for the four held-out validation specimens. Both datasets had MAE below \SI{5}{\percent} at {$N=0$} and {$N=15$}.

% Queue the plot before the final results paragraph so they can share a page.
\begin{figure*}[!t]
\begin{adjustbox}{width=1.0\linewidth,center}
  \input{Images/bainite_predictions_panel.pgf}
    \end{adjustbox}
  \caption{Predicted versus dilatometry-derived reference bainite fractions for all datasets with {$N=0$} (left) and {$N=15$} (right). Each point is the mean of predictions from 15 sub-images per source image included in inference, pooled over one reference fraction. {Error bars show the standard deviation of these pooled sub-image predictions.} Lines show descriptive linear fits.}
  \label{fig:bainite_predictions_panel}
\par\vspace{12pt}
\begin{minipage}{\textwidth}
\centering
    \captionsetup{labelfont={color=black}}
    \captionof{table}{Descriptive metrics computed from equally weighted mean predictions for each reference bainite fraction. MAE is expressed in \%, MSE in \%$^2$, and $R^2$ is dimensionless. The baseline predicts a constant \SI{50}{\percent}, the mean of the $\text{M}_1^\mathrm{Train}$ reference fractions.}
    \label{tab:metrics_meanpred}
    \normalsize
    \begin{tabular}{lrrrrrrrrr}
            \toprule
            Dataset & \multicolumn{3}{c}{{$N=0$}} & \multicolumn{3}{c}{{$N=15$}} & \multicolumn{3}{c}{Constant-50 baseline} \\
             & MSE & MAE & $R^2$ & MSE & MAE & $R^2$ & MSE & MAE & $R^2$ \\
            \midrule
            $\text{M}_1^\text{Train}$ & 4.38 & 1.52 & 0.99 & 11.73 & 2.73 & 0.99 & 816.67 & 24.67 & 0.00 \\
            $\text{M}_1^\text{Validation}$ & 18.90 & 4.13 & 0.96 & 26.55 & 4.51 & 0.95 & 500.00 & 20.00 & 0.00 \\
            $\text{M}_1^\text{Reprep}$ & 78.83 & 6.94 & 0.88 & 69.67 & 6.61 & 0.90 & 666.67 & 22.22 & 0.00 \\
            $\text{M}_2$ & 1843.95 & 35.70 & -1.24 & 354.74 & 14.76 & 0.57 & 825.00 & 25.00 & 0.00 \\
            $\text{M}_3$ & 625.40 & 18.22 & 0.06 & 393.04 & 13.84 & 0.41 & 666.67 & 22.22 & 0.00 \\
            \bottomrule
    \end{tabular}
\end{minipage}
\end{figure*}

For $\text{M}_1^\text{Reprep}$, predictions retained good agreement with the reference fractions after repeated preparation, with MAE values of 6.94 and 6.61 percentage points and corresponding $R^2$ values of 0.88 and 0.90 at {$N=0$} and {$N=15$}, respectively. For $\text{M}_2$ and $\text{M}_3$, {$N=15$} reduced MAE to \SI{14.76}{\percent} and \SI{13.84}{\percent}, respectively. These errors were lower than the corresponding constant baselines but remained substantially higher than for the $\text{M}_1$ datasets.

% Keep the complete results figure sequence before the discussion.
\clearpage
% Restore the text-column height after flushing the double-column floats.
\global\vsize=\textheight

\section{Discussion}

The results show that optical micrographs contain information useful for estimating bainite fraction even when bainitic and martensitic regions cannot be reliably delineated by visual inspection. Direct regression addresses a different task from constituent segmentation: it estimates a fraction from the overall image appearance without requiring explicit boundaries between constituents. This reduces the annotation requirements and is relevant to {bearing heat-treatment} applications where the quantity of interest is the bainite fraction rather than its spatial distribution.

The held-out validation results ($\text{M}_1^{\text{Validation}}$) at {$N=0$} demonstrate interpolation to intermediate bainite fractions within the same microscope domain without {the iterated mean filter}. Predictions remain in close agreement with the reference fractions after repeated preparation ($\text{M}_1^{\text{Reprep}}$), despite an increase in error that indicates some sensitivity to surface preparation and spatial sampling within the original specimens. The larger errors for the additional microscopes ($\text{M}_2$ and $\text{M}_3$) show reduced transferability across imaging workflows. These comparisons encompass differences in {metallographic preparation,} objectives, microscope platforms, acquisition software, export settings, and procedures.

The differences in contrast and apparent sharpness in Figure~\ref{fig:preprocessing_histograms} illustrate variability across the evaluated preparation and imaging workflows. Visual appearance alone does not establish how these differences affect reference-fraction prediction, which is assessed quantitatively against the dilatometry-derived reference fractions. The cross-microscope comparison therefore evaluates transfer between the complete workflows rather than isolating the influence of microscope hardware.

The good agreement observed after repeated metallographic preparation is encouraging for practical microscopy workflows, in which the exact surface condition and sampled fields cannot be reproduced identically. Predictions remained informative after grinding, polishing, and etching the specimens again and imaging different surface locations. Within the evaluated specimens, this indicates that useful predictive information persists across changes in preparation and spatial sampling.

The comb-like modulation in the $\text{M}_1$ histograms reflects the application of display settings during ZEN core export and the resulting nonuniform occupancy of the available 8-bit intensity levels. This configuration is relevant to operator workflows in which display settings are adjusted for visual assessment and the displayed appearance is retained on export. Such images may subsequently be used for automated analysis, making export settings relevant to transfer between imaging workflows.

{The iterated mean filter} suppresses the intensity pattern in the $\text{M}_1$ examples and also modifies local contrast and fine-scale image variation. Because all models were trained on $\text{M}_1$, the lower errors on $\text{M}_2$ and $\text{M}_3$ after the application of {the iterated mean filter} may reflect reduced sensitivity to the $\text{M}_1$ export mapping, to other differences between workflows, or to their combination. {The smoothing parameter value selected using $\text{M}_2$ alone also minimised the logged sub-image MSE on $\text{M}_3$, supporting the benefit of smoothing for both evaluated workflows.}

Future work will focus on domain adaptation to improve bainite-fraction predictions from images acquired with other microscopes. The aim is to reduce the influence of differences in image appearance between microscopy workflows and improve transferability beyond the training domain.

%%%%%%%%%%%%%%%%%%%%%%%%%%%%%%%%%%%%%%%%%%%%%%%%
%%%%%%%%%%%%%%%%%%%%%%%%%%%%%%%%%%%%%%%%%%%%%%%%
%%%%%%%%%%%%%%%%%%%%%%%%%%%%%%%%%%%%%%%%%%%%%%%%

\section{Conclusions}

A modified ResNet-18 was trained to predict dilatometry-derived reference bainite fractions directly from unsegmented optical micrographs of high-carbon bearing steel, without manual phase annotation or EBSD maps. The approach provides a route to fraction estimation {for bearing quality assessment} when overlapping optical features make explicit delineation of bainitic and martensitic regions difficult. Without {the iterated mean filter}, the mean predictions for four held-out validation specimens at intermediate reference fractions achieved a low mean absolute error, demonstrating interpolation within the same microscope domain.

Good predictive agreement was maintained after repeated metallographic preparation and sampling of different surface locations within the original specimens. These results indicate that useful predictive information persists after renewed grinding, polishing, and etching, supporting the potential of the method for microscopy workflows in which surface conditions and sampled fields vary.

Images acquired with additional microscopes presented a greater challenge. {The iterated mean filter} reduced the mean absolute errors for the two additional microscopes, respectively, showing its benefit in the evaluated workflows. The remaining differences in performance highlight the importance of preparation, acquisition, and export settings when transferring the model between microscopy workflows. Future work will focus on domain adaptation to reduce sensitivity to these differences and improve bainite-fraction predictions from images acquired with other microscopes.

% Submission TODO: add data and code availability, corresponding-author
% details, funding, author contributions, and competing-interest statements.

%%%%%%%%%%%%%%%%%%%%%%%%%%%%%%%%%%%%%%%%%%%%%%%%
%%%%%%%%%%%%%%%%%%%%%%%%%%%%%%%%%%%%%%%%%%%%%%%%
%%%%%%%%%%%%%%%%%%%%%%%%%%%%%%%%%%%%%%%%%%%%%%%%

\begingroup
\small
\RaggedRight
\setlength{\bibsep}{2pt}

\endgroup

\end{document}